\documentclass[11pt,a4paper]{article}
\usepackage[hyperref]{acl2019}
\usepackage{times}
\usepackage{latexsym}

\usepackage{todonotes}
\usepackage{url}

\usepackage{xr}

\makeatletter
\newcommand*{\addFileDependency}[1]{
  \typeout{(#1)}
  \@addtofilelist{#1}
  \IfFileExists{#1}{}{\typeout{No file #1.}}
}
\makeatother
 
\newcommand*{\myexternaldocument}[1]{%
    \externaldocument{#1}%
    \addFileDependency{#1.tex}%
    \addFileDependency{#1.aux}%
}

\myexternaldocument{appendix}

\aclfinalcopy 

\title{\textsc{MinWikiSplit}: A Sentence Splitting Corpus with Minimal Propositions}

\author{Christina Niklaus\textsuperscript{1}\textsuperscript{3}, Andr\'{e} Freitas\textsuperscript{2}, \and Siegfried Handschuh\textsuperscript{1}\textsuperscript{3} \\
  \textsuperscript{1} University of St.Gallen\\
  {\small{{\tt \{christina.niklaus, siegfried.handschuh\}{\tt @unisg.ch}}}}\\
  \textsuperscript{2} University of Manchester\\
  {\small{{\tt andre.freitas@manchester.ac.uk}}}\\
  \textsuperscript{3} University of Passau\\
  {\small{{\tt \{christina.niklaus, siegfried.handschuh\}{\tt @uni-passau.de}}}}
\\}

\date{}

\begin{document}
\maketitle
\begin{abstract}
    We compiled a new sentence splitting corpus that is composed of 203K pairs of aligned complex source and simplified target sentences. Contrary to previously proposed text simplification corpora, which contain only a small number of split examples,  
    we present a dataset where each input sentence is broken down into a set of minimal propositions, i.e. a sequence of sound, self-contained utterances with each of them presenting a minimal semantic unit that cannot be further decomposed into meaningful propositions.
    This corpus is useful for developing sentence splitting approaches that learn how to transform sentences with a complex linguistic structure into a fine-grained representation of short sentences that present a simple and more regular structure which is easier to process for downstream applications and thus facilitates and improves their performance. 
    
    
\end{abstract}

\section{Introduction}

Sentences that present a complex linguistic structure can be \textit{hard to comprehend by human readers}, as well as \textit{difficult to analyze by semantic applications} \cite{Saggion2017AutomaticTS}.
Identifying such grammatical complexities in a sentence and transforming them into simpler structures, using a set of text-to-text rewriting operations,
is the goal of syntactic text simplification (TS). One of the major types of operations that are used to perform this rewriting step is \textit{sentence splitting}: it divides a sentence into several shorter components, with each of them presenting a simpler and more regular structure that is easier to process by both humans and machines (see Table~\ref{tab:example}).

\begin{table*}[!ht]
\centering
  \begin{tabular}{ | l | p{13cm} | }
    \hline
    
    Complex source & The house was once part of a plantation and it was the home of Josiah Henson, a slave who escaped to Canada in 1830 and wrote the story of his life. \\ \hline
    
    \textsc{Min\-Wiki\-Split} & \textbf{The house was once part of a plantation. It was the home of Josiah Henson. Josiah Henson was a slave. This slave escaped to Canada. This was in 1830. This slave wrote the story of his life.} \\ \hline\hline
    
    Complex source & Gary Goddard, founder of Gary Goddard Entertainment, a company that designs theme parks, attractions and upscale resorts, estimated that about half his work in the last few years has been in Asia and the Middle East. \\ \hline
    
    \textsc{Min\-Wiki\-Split} & \textbf{About half his work in the last few years has been in Asia. This was what Goddard estimated. About half his work in the last few years has been in the Middle East. This was what Gary Goddard estimated. Gary Goddard was founder of Gary Goddard Entertainment. Gary Goddard Entertainment was a company. This company designs theme parks. This company designs attractions. This company designs upscale resorts.} \\ \hline\hline
    
    

 Complex source & The film is a partly fictionalized presentation of the tragedy that occurred in Kasaragod District of Kerala in India, as a result of endosulfan, a pesticide used on cashew plantations owned by the government. \\ \hline
 
\textsc{Min\-Wiki\-Split} &  \textbf{The film is a partly fictionalized presentation of the tragedy. This tragedy occurred in Kasaragod District of Kerala in India. This was as a result of endosulfan. Endosulfan is a pesticide. This pesticide is used on cashew plantations. These cashew plantations are owned by the government.}  \\ \hline
    
  \end{tabular} 
  
  \caption{Example instances from \textsc{MinWiki\-Split}. A complex source sentence is broken down into a set of syntactically simplified sentences by decomposing clausal and phrasal elements and transforming them into self-contained propositions that present a simple structure in the form of a sequence of subject, predicate, and optionally an object, adverbial or complement (or simple combination thereof).}
  \label{tab:example}
\end{table*} 

Syntactic TS with a focus on the task of sentence splitting has been attracting growing interest in the natural language processing (NLP) community within the past few years. One line of work targets reader populations with reading difficulties, such as people suffering from dyslexia, aphasia or deafness \cite{Siddharthan2014,Saggion:2015:MSI:2775084.2738046,Ferres2016}, 
 while 
the second line of work aims at generating an intermediate representation that is easier to process for downstream semantic tasks whose predictive quality deteriorates with sentence length and complexity. Prior work has established that applying syntactic TS as a preprocessing step can improve the performance of a variety of applications, including Machine Translation \cite{stajner2016can,stajner2018improvingMT}, Open Information Extraction \cite{cetto2018graphene}, or Text Summarization \cite{siddharthan2004syntactic,bouayad2009improving}.




\section{Limitations of Existing Sentence Splitting Corpora}
\label{sec:limitations}

All of the TS approaches mentioned above make use of a set of \textit{hand-crafted transformation rules} to decompose complex sentences into a sequence of structurally simplified components, requiring a complex rule engineering process. 
To overcome this expensive manual effort, \newcite{Narayan2017} presented a first attempt at modelling a data-driven sentence splitting approach where simplification rewrites are learned automatically from examples of aligned complex source and simplified target sentences. 
Since previously compiled TS corpora (PWKP \cite{zhu2010monolingual}, EW-SEW \cite{Coster:2011:SEW:2002736.2002865}, and Newsela \cite{Xu2015newsela}) contain only a small number of split examples, they are ill-suited for learning to decompose sentences into shorter, syntactically simplified components. Therefore, \newcite{Narayan2017} gathered a new dataset, \textsc{WebSplit}, which is the first TS corpus that explicitly addresses the task of sentence splitting, while abstracting away from deletion-based and lexical 
simplification operations. It is composed of over one million tuples that map a single complex sentence to a sequence of structurally simplified sentences. 

\newcite{aharoni2018split} criticized the data split proposed by \newcite{Narayan2017}. They observed that 99\% of the simple sentences (which make up for more than 89\% of the unique ones) contained in the validation and test sets also appear in the training set. Consequently, instead of learning to split and rephrase complex sentences, models that are trained on this dataset will be prone to learn to memorize entity-fact pairs. Hence, this split is not well suited for measuring a model's ability to generalize to unseen input sentences. To fix this issue, \newcite{aharoni2018split} present a new train-development-test data split where nearly no simple sentence that is contained in the development or test set occurs verbatim in the training set.

Lately, \newcite{Botha2018} discovered that the sentences from the \textsc{WebSplit} corpus contain fairly \textit{unnatural linguistic expressions} over only a \textit{small vocabulary} and a rather \textit{uniform sentence structure}, which is predominantly composed of a sequence of coordinate clauses, occasionally augmented with a relative or adverbial clause (see Table~\ref{tab:websplit_example}). To overcome these limitations, they present \textsc{Wiki\-Split}, a dataset of one million sentences that were mined 
from Wikipedia edit histories. This corpus provides a rich and varied vocabulary over naturally expressed sentences showing a diverse linguistic structure, and their extracted splits. 
However, there is only a single split per source sentence in the training set (see Table~\ref{tab:wikisplit_example}). Accordingly, when a model is trained on this dataset, it is susceptible to exhibiting a \textit{strong conservatism}, splitting each input sentence into exactly two output sentences only. Consequently, the resulting simplified sentences are still comparatively long and complex, mixing multiple, potentially semantically unrelated propositions that are difficult to analyze for downstream tasks. 

\begin{table}[!ht]
\centering
  \begin{tabular}{ | p{7.2cm} | }
  \hline
  (1) A Loyal Character Dancer was published by Soho Press, in the United States, where some Native Americans live. \\
  (2) Dead Man's Plack is in England and one of the ethnic groups found in England is the British Arabs.\\
  
    \hline
  \end{tabular} 
  
  \caption{Characteristic example source sentences from the \textsc{WebSplit} corpus.}
  \label{tab:websplit_example}
\end{table} 

\begin{table}[!ht]
\centering
  \begin{tabular}{ | p{1.6cm} | p{5.2cm} | }
  \hline
  Complex source & Starring Meryl Streep, Bruce Willis, Goldie Hawn and Isabella Rossellini, the film focuses on a childish pair of rivals who drink a magic potion that promises eternal youth. \\ \hline
  
  Simplified output & Starring Meryl Streep, Bruce Willis, Goldie Hawn and Isabella Rossellini. The film focuses on a childish pair of rivals who drink a magic potion that promises eternal youth. \\ \hline\hline
  
  Complex source & The Assistant Attorney in Orlando investigated the modeling company, and decided that they were not doing anything wrong, and after Pearlman's bankruptcy, the company emerged unscathed and was sold to a Canadian company. \\ \hline
  
  Simplified output & The Assistant Attorney in Orlando investigated the modeling company, and decided that they were not doing anything wrong. After Pearlman's bankruptcy, the modeling company emerged unscathed and was sold to a Canadian company. \\ \hline
  
  
  \end{tabular} 
  
  \caption{Pairs of aligned complex source and simplified target sentences from \textsc{Wiki\-Split}.}
  \label{tab:wikisplit_example}
\end{table}

\section{\textsc{MinWiki\-Split} Corpus} 





We improve on previously compiled sentence splitting corpora and present \textsc{MinWiki\-Split},\footnote{The \textsc{MinWiki\-Split} dataset is publicly released under \url{https://github.com/Lambda-3/MinWikiSplit}.} a new dataset that can be used to train models for the task of decomposing sentences with a complex linguistic structure into a simplified representation that presents a more regular structure which is easier to process for downstream semantic applications and may support a faster generalization in machine learning tasks. This output may serve as an intermediate representation to facilitate and improve the performance of a wide range of artificial intelligence (AI) tasks.

Since shorter sentences are generally better processed by NLP systems \cite{Narayan2017}, we aimed at gathering a corpus where \textbf{each complex source sentence is broken down into a set of minimal propositions}, i.e. a sequence of sound, self-contained utterances, with each of them presenting a minimal semantic unit that cannot be further decomposed into meaningful propositions \cite{bast2013open}. Thus, we augment the Split-and-Rephrase task that was originally defined in \newcite{Narayan2017} by the notion of \textit{minimality}. In that way, we intend to overcome the conservatism exhibited by state-of-the-art structural TS approaches, which tend to retain the input rather than transforming it, and expect to improve the performance of a wide range of AI tasks. 

\section{Corpus Construction}
\textsc{MinWiki\-Split} is a large-scale sentence splitting corpus consisting of 203K complex source sentences and their simplified counterparts in the form of a sequence of minimal propositions. It was created by running \textsc{DisSim} \cite{niklaus-etal-2019-transforming}, a syntactic TS framework, over the one million complex input sentences from the \textsc{Wiki\-Split} corpus. \textsc{DisSim} applies a small set of 35 hand-written transformation rules to decompose a wide range of linguistic constructs, including both clausal components (coordinations, adverbial clauses, relative clauses and reported speech) and phrasal elements (appositions, prepositional phrases, adverbial/adjectival phrases and coordinate noun phrases). In that way, a fine-grained output in the form of a sequence of minimal, self-contained propositions is produced. Some example instances are shown in Table~\ref{tab:example}. 


\begin{table*}[!ht]
\centering
  \begin{tabular}{ |l  || c | c | c | c | c | c | }
    \hline
    & \#T/S & \#S/C & \%\newline SAME & LD\textsubscript{SC} & SAM\-SA & SAM\-SA\textsubscript{abl}  \\ \hline \hline
    Complex & 30.75 & 1.18 & 100 & 0.00  & 0.36 &  0.94  \\ \hline
    \textsc{MinWikiSplit} & 12.12 & 3.84 & 0.00 & 17.73 & 0.40 & 0.48  \\ \hline
    
  \end{tabular} 
  
  \caption{Results of the automatic evaluation procedure on a random sample of 1000 sentences.} 
  \label{results_automatic}
\end{table*}

To ensure that the resulting dataset is of high quality, we defined a set of dependency parse and part of speech based heuristics 
to filter out sequences that contain \textit{grammatically incorrect} sentences, as well as sentences that \textit{mix multiple semantic units} and, thus, are violating the specified minimality requirement. For instance, in order to verify that the simplified sentences are grammatically sound, we check whether the root node of the output sentence is a verb and whether one of its child nodes is assigned a dependency label that denotes a subject component.
To test if the simplified sentences represent minimal propositions, we check whether the output does not contain a clausal modifier, such as a relative clause modifier, adverbial clause modifier or a clausal modifier of a noun. Moreover, we ensure that no conjunction is included in the simplified output sentences. 
In the future, we will implement some further heuristics to \textit{avoid uniformity} in the structure of the source sentences. In that way, we aim at guaranteeing a great structural variability in the input in order to enable systems that are trained on the \textsc{MinWiki\-Split} corpus to learn splitting rewrites for a wide range of linguistic constructs.



After running the sentence simplification framework \textsc{DisSim} over the sentences from the \textsc{Wiki\-Split} corpus and applying the set of heuristics that we defined to ensure grammaticality and minimality of the output, 203K pairs of input and corresponding output sequences were left.




\section{Experiments}
We performed both a manual analysis and an automatic evaluation to assess the quality of the produced corpus.

\subsection{Automatic Metrics}
To estimate the quality of the simplified target sentences of the \textsc{MinWiki\-Split} corpus, we computed some basic statistics, including (i) the average sentence length of the simplified sentences in terms of the average number of tokens per output sentence (\#T/S); (ii) the average number of simplified output sentences per complex input (\#S/C); (iii) the percentage of sentences that are copied from the source without performing any simplification operation (\%SAME), serving as an indicator for conservatism, i.e. the tendency to retain the input rather than transforming it; and (iv) the averaged word-based Levenshtein distance from the input (LD\textsubscript{SC}), which provides further evidence for how reluctant the underlying system is in splitting the input into minimal semantic units.

Moreover, to measure the structural simplicity of the instances contained in \textsc{Min\-Wiki\-Split}, we calculated the SAMSA and SAMSA\textsubscript{abl} scores of both the complex source and the simplified output sentences \cite{sulemsemantic}. They are the first metrics that explicitly target syntactic aspects of TS. The SAMSA metric is based on the idea that an optimal split of the input is one where each predicate-argument structure is assigned its own sentence in the simplified output and measures to what extent this assertion holds for the input-output pair under consideration. Accordingly, the SAMSA score is maximized when each split sentence represents exactly one semantic unit in the input. SAMSA\textsubscript{abl} does not penalize cases where the number of sentences in the simplified output is lower than the number of events contained in the input, indicating separate semantic units that should be split into individual target sentences for obtaining minimal propositions.\footnote{Prior work on syntactic TS commonly also reports average BLEU \cite{papineni2002bleu} scores. However, \newcite{sulemBLEU2018} recently demonstrated that this score is inappropriate for the evaluation of TS approaches when sentence splitting is involved. Therefore, we refrain from calculating BLEU scores.}

These computations were carried out on a random sample of 1000 sentences from \textsc{Min\-Wiki\-Split}. The results are provided in Table \ref{results_automatic}. The scores demonstrate that on average our proposed sentence splitting corpus contains four simplified target sentences per complex source sentence, with every target proposition consisting of 12 tokens. Moreover, no input is simply copied to the output, but split into smaller components. Both the high averaged Levenshtein distance of almost 18 and the SAMSA score (0.40) confirm previous findings. The latter is highly correlated with structural simplicity and grammaticality, indicating that the output sentences contained in our corpus are grammatically sound and present a simpler syntax than the input. With 0.48, we reach a decent score for the simplified target sentences with regard to SAMSA\textsubscript{abl}, too, which has a high correlation with meaning preservation. 

\subsection{Manual Analysis}\label{sec:manual_analysis}
In a second step, we randomly selected a subset of 300 sentences from \textsc{MinWiki\-Split}, on which we conducted a manual analysis in order to get some deeper insights into the quality of the simplified sentences.
Each input-output pair was rated by 2 non-native, but fluent English speakers according to three parameters: grammaticality, meaning preservation and structural simplicity (see Table \ref{tab:human_eval}).

\begin{table}[!ht]
\centering
  \begin{tabular}{ | l | p{6.5cm} | }
    \hline
    G & Is the output fluent and grammatical? \\\hline
    M & Does the output preserve the meaning of the input? \\\hline
    S & Is the output simpler than the input, ignoring the complexity of the words? \\
    \hline
     \end{tabular} 
  
  \caption{Questions for the human evaluation.}
  \label{tab:human_eval}
\end{table}

The inter-annotator agreement was computed using Cohen’s quadratic weighted $\kappa$ \cite{cohen1968weighted}. The obtained rates were 0.24, 0.25 and 0.75 for grammaticality, meaning preservation and structural simplicity, respectively. System scores were calculated by averaging over the annotators' scores and the 300 sentences. 

\begin{table}[!ht]
\centering
  \begin{tabular}{ | c | c | c |}
    \hline
     G & M & S  \\ \hline 
      \textbf{4.36}  & \textbf{4.10} & \textbf{3.43}  \\
      \hline
  
     \end{tabular} 
  
  \caption{Averaged human evaluation ratings on a random sample of 300 sentences from \textsc{MinWiki\-Split}. Grammaticality (G), meaning preservation (M) and structural simplicity (S) are measured using a 1 (very bad) to 5 (very good) scale.} 
  \label{resultsHumanEval}
\end{table}

The results of the human evaluation are displayed in Table \ref{resultsHumanEval}.
These scores show that we succeed in producing output sequences that reach a high level of grammatical soundness and almost always perfectly preserve the original meaning of the input. The third dimension under consideration, structural simplicity, which captures the degree of minimality in the simplified sentences, scores high values, too. However, our manual analysis revealed some room for improvement. Consequently, in the future, we plan to implement stricter heuristics for sorting out output sequences that still mix multiple semantically unrelated propositions.



\section{Conclusion}
We compiled \textsc{MinWiki\-Split}, a sentence splitting corpus consisting of 203K complex source sentences and their split counterparts. This dataset can be used to train natural language generation applications that perform a syntactic TS, simplifying sentences with a complex linguistic structure into a fine-grained representation of short sentences that present a simple and more regular structure. The thus generated output may serve as an intermediate representation that is easier to process for downstream semantic applications and may thus lead to a better performance of those tools. We intend to train a sentence simplification model on \textsc{MinWikiSplit} and compare it to previously proposed systems trained on the \textsc{WebSplit} and \textsc{WikiSplit} corpora.

Moreover, we plan to improve the quality of the simplified target sentences in our corpus in accordance with the insights we gained through the analyses described above. First of all, we will perform a detailed error analysis of the output to determine the most common types of mistakes and get some starting points for further improving our heuristics for filtering out malformed simplifications. To enhance the syntactic correctness of the output, we will train a classifier on the recently proposed CoLA dataset \cite{warstadt2018neural} to eliminate instances with ungrammatical target sentences from our corpus. In addition, special attention will be given to improving the heuristics that ensure that each simplified target sentence represents a single semantic unit.

\bibliography{acl2019}
\bibliographystyle{acl_natbib}

\end{document}